\documentclass[10pt,twocolumn,letterpaper]{article}

\usepackage[review,algorithms]{wacv}      

\usepackage[T1]{fontenc}
\usepackage{amsmath}
\usepackage{multirow}
\usepackage{pbox}
\usepackage{utfsym}
\usepackage[shortlabels]{enumitem}
\usepackage{amssymb}

\usepackage{multirow}
\usepackage{multicol}
\usepackage{pbox}
\usepackage{amsmath}
\usepackage{enumitem}
\usepackage{amsmath, amssymb, amsfonts}
\usepackage{pifont}
\usepackage{comment}

\usepackage{graphicx}
\usepackage{amssymb}
\usepackage{booktabs}
\usepackage{enumitem}
\usepackage{amssymb}
\usepackage{amsmath}
\usepackage{bm}
\usepackage{multirow}
\usepackage{pbox}
\usepackage{graphicx}
\usepackage{xfakebold}
\newcommand{\fbseries}{\unskip\setBold\aftergroup\unsetBold\aftergroup\ignorespaces}
\makeatletter
\newcommand{\setBoldness}[1]{\def\fake@bold{#1}}
\makeatother
\usepackage[pagebackref,breaklinks,colorlinks]{hyperref}

\usepackage[capitalize]{cleveref}
\crefname{section}{Sec.}{Secs.}
\Crefname{section}{Section}{Sections}
\Crefname{table}{Table}{Tables}
\crefname{table}{Tab.}{Tabs.}

\def\confName{WACV}
\def\confYear{2025}

\newcommand\figref{Figure~\ref}
\newcommand\tabref{Table~\ref}

\newcommand{\cmark}{\ding{51}}%
\newcommand{\xmark}{\ding{55}}%

\begin{document}

\title{\textnormal{This paper has been accepted for WACV 2025} \\[1ex]
CAMS: Convolution and Attention-Free Mamba-based Cardiac Image Segmentation}

\author{
    Abbas Khan\\
    School of Electronic Engineering and Computer Science, Queen Mary University of London, UK\\
    Queen Mary's Digital Environment Research Institute (DERI), London, UK\\
    {\tt\small acw676@qmul.ac.uk}
    \and
    Muhammad Asad\\
    School of Biomedical Engineering and Imaging Sciences King’s College London, UK\\
    Queen Mary's Digital Environment Research Institute (DERI), London, UK\\
    {\tt\small muhammad.asad@qmul.ac.uk}
    \and
     Martin Benning\\
     Department of Computer Science, University College London, UK \\
    Queen Mary's Digital Environment Research Institute (DERI), London, UK\\
    {\tt\small martin.benning@ucl.ac.uk}
    \and
    Caroline Roney\\
    School of Engineering and Materials Science, Queen Mary University of London, UK\\
    Queen Mary's Digital Environment Research Institute (DERI), London, UK\\
    {\tt\small c.roney@qmul.ac.uk}
    \and
    Gregory Slabaugh\\
    School of Electronic Engineering and Computer Science, Queen Mary University of London, UK\\
    Queen Mary's Digital Environment Research Institute (DERI), London, UK\\
    {\tt\small g.slabaugh@qmul.ac.uk}
}

\maketitle

\begin{abstract}
    Convolutional Neural Networks (CNNs) and Transformer-based self-attention models have become the standard for medical image segmentation. This paper demonstrates that convolution and self-attention, while widely used, are not the only effective methods for segmentation. Breaking with convention, we present a \textbf{C}onvolution and self-\textbf{A}ttention-free \textbf{M}amba-based semantic \textbf{S}egmentation \textbf{Net}work named CAMS-Net. Specifically, we design Mamba-based \emph{Channel Aggregator} and \emph{Spatial Aggregator}, which are applied independently in each encoder-decoder stage. The Channel Aggregator extracts information across different channels, and the Spatial Aggregator learns features across different spatial locations. We also propose a \textbf{L}inearly \textbf{I}nterconnected \textbf{F}actorized \textbf{M}amba (LIFM) block to reduce the computational complexity of a Mamba block and to enhance its decision function by introducing a non-linearity between two factorized Mamba blocks. Our model outperforms the existing state-of-the-art CNN, self-attention, and Mamba-based methods on CMR and M\&Ms-2 Cardiac segmentation datasets, showing how this innovative, convolution, and self-attention-free method can inspire further research beyond CNN and Transformer paradigms, achieving linear complexity and reducing the number of parameters. Source code and pre-trained models are available at: https://github.com/kabbas570/CAMS-Net.
\end{abstract}

\section{Introduction}
\label{sec:intro}
Image segmentation is an essential part of Cardiac image analysis \cite{martin2020image}. It can help quantify the size and shape of different regions of interest, such as left ventricle (LV), right ventricle (RV), left atrium (LA), right atrium (RA), myocardium (MYO), useful for monitoring disease progression, prognosis and supporting computer-aided intervention \cite{chen2020deep}. Manual segmentation is considered a gold standard; however, with the developments in artificial intelligence, recent research has been focused on developing automatic methods for faster,  cheaper, and reproducible results \cite{litjens2019state,hesamian2019deep}.

Convolutional neural networks (CNNs) and transformer-based self-attention mechanisms have significantly evolved the landscape of medical image segmentation \cite{yao2024cnn}. Although CNNs have been the commonly used choice \cite{aljuaid2022survey}, current literature suggests that self-attention-based methods produce better results than CNN architectures \cite{khan2023transformers} due to their global receptive field, ability to model long-range dependencies, and the dynamic weights mechanism \cite{dosovitskiy2020image}. CNN-based architectures have been criticized for their limited receptive field \cite{richter2021should}, their limited ability to effectively capture long-range dependencies, and their bias toward recognizing textures rather than shapes \cite{geirhos2018imagenet}. However, attention-based methods are computationally expensive compared to CNNs due to their quadratic complexity \cite{niu2021review} and excessive memory requirements \cite{ramachandran2019stand}. Recent research work has focused on reducing the computational complexity of attention-based methods while maintaining accuracy, including efficient additive attention \cite{shaker2023swiftformer}, efficient self-attention \cite{gao2021utnet}, and separable self-attention \cite{mehta2022separable}. Hybrid CNN-transformer-based segmentation methods have also been a recent trend that harnesses relative strengths of CNN and self-attention \cite{tang2024htc,khan2024crop}. These methods combine CNN and self-attention layers to capture local and global features while reducing self-attention's computational complexity.

Recently, Mamba has gained prominence in the computer vision field and integrates Gated MLP \cite{mehta2022long} into the State Space Model (SSM) of H3 \cite{dao2023hungry}. Readers are encouraged to refer to \cite{patro2024mamba,zhang2024survey} for a more comprehensive understanding of this topic. SSMs \cite{gu2021efficiently} such as Mamba \cite{gu2023mamba} 
are considered as a potential replacement for transformers because they can capture long-range dependencies while maintaining linear computation complexity. Several architectures have been proposed to show the power of Mamba for computer vision tasks, including Vision Mamba \cite{zhu2024vision}, Visual Mamba (VMamba) \cite{liu2024vmamba},  ZigMa \cite{hu2024zigma}, and medical image segmentation is no exception \cite{ruan2024vm,wang2024mamba}. The majority of existing encoder-decoder medical image segmentation architectures inspired by Mamba \cite{gu2023mamba} and vision transformers \cite{dosovitskiy2020image} differ in how the convolution layers, self-attention, and Mamba blocks are arranged.

In this paper, we go beyond the arrangements of these blocks and propose a novel convolution and attention-free CAMS-Net for medical image segmentation. We propose Mamba-based spatial and channel aggregators to extract information across different channels and spatial locations, along with a Linearly Interconnected Factorized Mamba (LIFM) block to further reduce the computational complexity and enhance its decision function. Unlike CNN-based segmentation networks, it excels at capturing global features while surpassing self-attention-based methods by modeling long-range dependencies with linear rather than quadratic complexity. Compared to other Mamba-based segmentation networks, CAMS-Net stands out for its convolutional-free design, eliminating the need for hybrid architectures. Along with these innovations, the CAMS-Net outperforms existing networks, making it a more efficient and effective solution for medical image segmentation.

\section{Related Work}
UNet \cite{ronneberger2015u} is a pioneering network architecture for medical image segmentation, and several subsequent architectures, including ResUNet \cite{diakogiannis2020resunet}, UNet++ \cite{zhou2019unet++}, have extended its initial formulation. These networks use an encoder-decoder design, where the encoder extracts information from images, and the decoder reconstructs the segmentation map. Skip connections \cite{drozdzal2016importance} mitigate the vanishing gradient problem and reuse the features from the encoder side. With the advent of the vision transformer \cite{dosovitskiy2020image}, self-attention-based methods have become popular in medical image segmentation to overcome the limitations of CNN-based pipelines. These methods include Swin-UNet \cite{cao2022swin}, which replaces the convolutional layers with Swin-Transformer blocks \cite{liu2021swin}. The Swin-UNet also follows a U-shaped architecture, where the encoder utilizes a hierarchical Swin Transformer with shifted windows to extract context features and a symmetric decoder with patch-expanding layers for upsampling. The UNEt TRansformers (UNETR) \cite{hatamizadeh2022unetr} uses a transformer-based encoder to learn sequence representations of the input images, allowing the network to capture global multi-scale information and a  CNN-based decoder for localized information. 
Similar to  Swin-UNet \cite{cao2022swin}, Swin-UNETR \cite{hatamizadeh2021swin} is also built on a hierarchical Swin transformer. However, it has a hybrid architecture that only uses a Swin transformer in the encoder to extract features and a CNN-based decoder to generate the segmentation map.

 Mamba-UNet \cite{wang2024mamba} incorporates the VMamba-based \cite{liu2024vmamba} encoder-decoder structure with UNet. The Cross-Scan Module from VMamba scans the input image in four ways to integrate information from all other locations for each element of the features. Mamba-UNet utilizes these VMamba \cite{liu2024vmamba} blocks throughout the U-shaped architecture to capture semantic contexts from intensity images. 
Vision Mamba UNet (VM-UNet) \cite{ruan2024vm} extends Vision Mamba \cite{zhu2024vision} using foundation blocks named Visual State Space. Its asymmetrical encoder-decoder structure leverages the power of SSMs to capture contextual information while maintaining linear computational complexity.

Nevertheless, these hybrid methods address the challenges posed by self-attention and CNNs and utilize both local and global features in dense prediction tasks, like segmentation. However, convolution-free methods have become a recent trend in computer vision, and some of these approaches have tried to utilize self-attention-based architectures only. For example, Kim et al. \cite{kim2022restr} proposed ReSTR for referring image segmentation, where transformer-based encoders extract features from each modality, image, and text, followed by coarse-to-fine segmentation decoder transforms to reconstruct the output from fused features. Karimi et al. \cite{karimi2021convolution} proposed a convolution-free 3D network for medical image segmentation. The 3D image block is divided into $n^k$ patches ($k$=3, or 5) and computes a 1D embedding for each patch. Their method predicts the center patch of the block using self-attention between patch embeddings. 
MLP-Mixer \cite{tolstikhin2021mlp} proposed an alternative architecture for image classification tasks built solely on multi-layer perceptrons (MLPs). The channel- and token-mixing MLPs learn the per-location features and between different spatial locations (tokens), respectively.

Although these most recent models attempt to overcome challenges posed by CNNs, they are based on self-attention with quadratic computational complexity and high memory requirements. This paper introduces a convolution-free and self-attention-free model to mitigate the limitations of convolution-based and self-attention-based architectures while maintaining the benefits that self-attention brings, i.e., the global receptive field, dynamic weight mechanism, and long-range dependencies at the expense of linear complexity. The contributions of this work are:
\begin{enumerate}
    \item To the best of our knowledge, we are the first to propose a convolution and self-attention-free Mamba-based segmentation network, CAMS-Net.
    \item We propose a Linearly Interconnected Factorized Mamba (LIFM) block to reduce the trainable parameters of Mamba and improve its non-linearity. LIFM implements a weight-sharing strategy for different scanning directions, specifically for the two scanning direction strategies of vision Mamba \cite{zhu2024vision}, to reduce the computational complexity further whilst maintaining accuracy.
    \item We propose Mamba Channel Aggregator (MCA) and Mamba Spatial Aggregator (MSA) and demonstrate how they can learn information along the channel and spatial dimensions of the features, respectively.
    \item Extensive experimental validation, including ablation studies, are conducted to showcase the efficacy of our proposed model. Our proposed CAMS-Net outperforms existing state-of-the-art segmentation models on the CMR and the Multi-Disease, Multi-View, and Multi-Center (M\&Ms-2) segmentation datasets, including pure CNN, self-attention, and hybrid self-attention, as well as methods using the original Mamba-based architecture combined with CNNs. 
\end{enumerate}
\begin{figure*}[t!]
\centering
\includegraphics[scale=0.35]{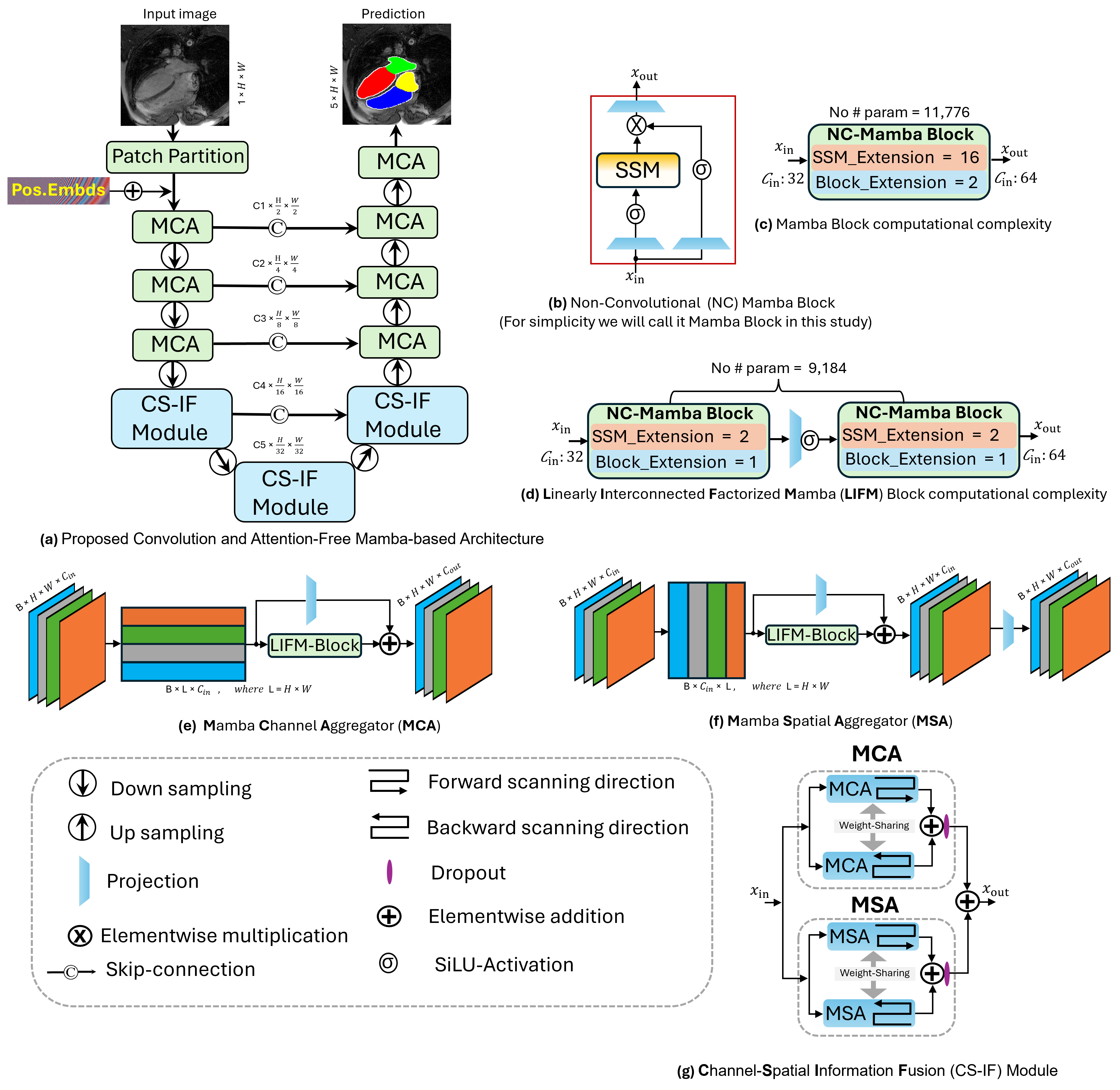}
\caption{(a) Overall architecture of proposed CAMS-Net, (b) the Non-Convolutional (NC) Mamba Block without local convolution, (c) comparison of proposed LIFM block with original Mamba, (e) Mamba Channel Aggregator (MCA), (f) Mamba Spatial Aggregator (MSA), and (g) the Channel-Spatial Information Fusion (CS-IF) Module.}
\label{p1}
\end{figure*}

\section{Methodology}
\label{sec:method}
The proposed convolution and a self-attention-free segmentation network, CAMS-Net, is shown in \figref{p1} (a). The input image is transformed into non-overlapping patches with a patch size of 2 $\times$ 2, reducing the in-plane spatial resolution by 2, and a linear embedding layer to project the features into dimension \emph{C1 = 64}. It also incorporates sinusoidal positional embeddings to encode spatial context information, enabling the encoder to understand the relative positions of different regions within the image. The features are also downsampled at each encoder's stage using a  2$\times$2 average pooling layer. In the next encoder stage and bottleneck, we implement the CS-IF module, allowing the model to learn richer features along channel and spatial dimensions.\\

On the decoder side, the features are upsampled at each stage using a bilinear interpolation window of 2$\times$2 to match the output dimension, followed by the CS-IF module in the first stage after the bottleneck and MCA in all other decoder stages. The skip connections \cite{drozdzal2016importance} are also implemented at each encoder-decoder stage to reuse the features and for faster convergence. 
Finally, a five-class segmentation map (one for each class, LA, RA, LV, RA, and background) 
is generated, followed by a Softmax activation. This section will explain the components of the CAMS-Net. 

\subsection{Factorized Mamba}
\label{sec:expF}
Inspired by deep convolutional neural networks \cite{simonyan2014very}, where a stack of two $3\times3$ convolution filters have an effective receptive field of $5\times5$, we propose the idea of factorized Mamba, which makes the decision function more discriminative and also reduces the number of parameters. The `Mamba block expansion factor' \emph{(E)} and the `SSM state expansion factor' \emph{(D)} control the overall complexity of the Mamba block. More specifically, \emph{E} expands the dimensions of the Mamba block using linear layers with learned weights $\mathbf{W}_{1}$, and $\mathbf{W}_{2}$, while \emph{D} projects the dimension within the SSM. We implemented the Mamba block with various \emph{E} and \emph{D} factors and analyzed their computational complexity, shown in Table 1 of the supplementary material. In the Mamba block, most of the parameters stem from \emph{E},  with minimal increment from \emph{D}.
The majority of the Mamba-based networks use the default SSM and Mamba block extension, shown in \figref{p1} (c), which is computationally expensive, and a single Mamba block brings 11,776 trainable parameters (for $c_{in}$=32 and $c_{out}$=64). Mathematically, it can be represented as,
\begin{equation}\label{eq1}
\mathbf{x}_{\text{out}} = \textbf{NC-MambaBlock}(\mathbf{x}_{\text{in}}, D, E),   
\end{equation}
with,
\begin{equation}\label{eq1}
\mathbf{x}_{\text{out}} = \mathbf{W}_{3} \Bigl( \sigma(\mathbf{W}_{2}\mathbf{x}_{in}) \odot \mathbf{SSM}(\sigma(\mathbf{W}_{1}\mathbf{x}_{in})) \Bigr),   
\end{equation}
where $\mathbf{W}_{1}$,$\mathbf{W}_{2}$,$\mathbf{W}_{3}$ are learnable weights for linear layers shown in \figref{p1} (b) used for input $x_{in}$ projection, $\odot$ represents element-wise multiplication  and $\sigma$ SiLU activation \cite{elfwing2018sigmoid}.

Our factorized Mamba block splits the extension parameters for SSM and Mamba, shown in \figref{p1} (d). We also add a linear layer followed by SiLU activation \cite{elfwing2018sigmoid} between two Mamba blocks to add more non-linearity and named it the Linearly Interconnected Factorized Mamba (LIFM) block, used throughout our proposed architecture. A single factorized Mamba block has 4,608 parameters (for $c_{in}$=32 and $c_{out}$=64), and the proposed LIFM-Block requires only 9,184 trainable parameters. For the first factorized Mamba block and linear layer, $C_{in}$ = $C_{out}$ = 32, and for the second factorized Mamba block, $C_{out}$ = 64.  Mathematically, we can represent LIFM Block as
\begin{eqnarray}\label{eq1}
\mathbf{x}_{\text{1}} & = & \textbf{NC-MambaBlock}(\mathbf{x}_{\text{in}}, D_{1}, E_{1}) \\
\mathbf{x}_{\text{2}} & = & \sigma(\mathbf{W}_{fm}\mathbf{x}_{\text{1}}) \\
 \mathbf{x}_{\text{out}} & = & \textbf{NC-MambaBlock}(\mathbf{x}_{\text{2}}, D_{2}, E_{2})
\end{eqnarray}
where, $D_{1}$ = $D_{2}$ = 2 , and $E_{1}$ = $E_{2}$ = 1, $\mathbf{W}_{fm}$ represents the linear layer between factorized two Mamba blocks. \\

Empirically, we also found that a large Mamba block can easily overfit the data and increase the overall computational burden of the network. So, we factorized the larger Mamba blocks at each stage and used two consecutive relatively smaller ones. This factorized approach reduces the number of trainable parameters and helps the network to increase its non-linearity to learn more complex patterns and representations in the data.
\subsection{Mamba Channel Aggregator}
\label{sec:CA}
The Mamba channel aggregator (MCA) aims to learn cross-channel information, as shown in \figref{p1} (e), learning the per-location features at different channels. Similar to a UNet structure, the number of channels is increased as $\{64,128,256,512,1024\}$ at each encoder stage and decreased as $\{512,256,128,64\}$ at each decoder stage. For the channel aggregator, the incoming features $\mathbb{R}^{B \times C \times H \times W}$ are reshaped to $\mathbb{R}^{B \times L \times C}$, where $L = H \times W$.
Then, the input is divided into two branches where, in one branch, the LIFM Block is applied, and the second branch acts as a residual connection, where a linear layer is used, followed by an element-wise addition operation with the features of the first branch. Mathematically, it can be represented as, 
\begin{equation}\label{eq1}
 x_{\text{out}} = \overline{\mathbf{f_{1}}}\Bigl(\mathbf{LIFM\_Block}\bigl(\mathbf{f_{1}}(x_{\text{in}})) \oplus \mathbf{W}_{c}(\mathbf{f_{1}}(x_{\text{in}}))\Bigl),
\end{equation}
where, $\mathbf{f}_{1}:\mathbb{R}^{B \times C \times H \times W} \rightarrow \mathbb{R}^{B \times L \times C}$ represents a reshaping function 
and $\overline{\mathbf{f_{1}}}: \mathbb{R}^{B \times L \times C} \rightarrow \mathbb{R}^{B \times C \times H \times W}$ performs the inverse operation, $\mathbf{W}_{c}$ is the residual linear layer of MCA, and $\oplus$ represents element-wise addition.
\subsection{Mamba Spatial Aggregator}
\label{sec:SA}
As shown in \figref{p1} (f), the Mamba spatial aggregator (MSA) aims to learn information about different spatial locations and enables communication amongst them. The spatial aggregator's computational complexity depends on the features' spatial dimensions, so it is only used for the lower-dimensional features of the U-shaped network. More specifically, it is used in the bottleneck, one encoder stage before the bottleneck, and one decoder stage after the bottleneck, shown in \figref{p1} (a). For the spatial aggregator, the incoming features $\mathbb{R}^{B \times C \times H \times W}$  are reshaped to $\mathbb{R}^{B \times C \times L}$. The features follow the same protocol as that of MCA, and finally, a linear layer is used either to expand (in the encoder) or to compress (in the decoder) the number of channels. In mathematical terms,  
\begin{equation}\label{eq1}
 \mathbf{x}_{\text{out}} = \mathbf{W}_{ci}\overline{\mathbf{f_{2}}}\Bigl(\mathbf{LIFM\_Block}\bigl(\mathbf{f_{2}}(\mathbf{x}_{\text{in}})\bigl) \oplus \mathbf{W}_{s}\bigl(\mathbf{f_{2}}(\mathbf{x}_{\text{in}})\bigl)\Bigl).
\end{equation}
Here, $\mathbf{f}_{2}:\mathbb{R}^{B \times C \times H \times W} \rightarrow \mathbb{R}^{B \times C \times L}$ represents a reshaping function 
and  $\overline{\mathbf{f_{2}}}: \mathbb{R}^{B \times C \times L} \rightarrow \mathbb{R}^{B \times C \times H \times W}$ does the inverse, $\mathbf{W}_{S}$ is the residual linear layer of MSA, and $\mathbf{W}_{ci}$ is a linear layer that either increases or decreases the number of channels in MSA, to match with MCA.

\subsection{Bidirectional Information Learning}
\label{sec:Bidir}
Inspired by Vision Mamba \cite{zhu2024vision}, we implemented both MCA and MSA using a bidirectional scanning arrangement scheme, shown in Figure 1 of the supplementary material. We incorporated the bidirectional SSMs to make the network spatially aware. Unlike Vision Mamba, we found that sharing the weights for two-direction schemes results in better average performance and also lowers computational complexity, as shown in the \tabref{tab1} of the ablation study. We also experimented with a multi-directional scanning arrangement, such as a four-directional \cite{liu2024vmamba} and an eight-directional scheme \cite{hu2024zigma}. However, the bidirectional scanning scheme augmented with the proposed weight-sharing strategy is the best practice for the task at hand due to the smaller dataset and the method's reduced complexity. 

\subsection{Channel-Spatial Information Fusion Module}
\label{sec:blck}
The Channel-Spatial Information Fusion (CS-IF) module comprises the MCA and MSA and merges the information extracted along channel and spatial dimensions, depicted in \figref{p1} (g). The incoming features are passed to the MCA and MSA, where each aggregation learns the features in both the forward and backward scanning directions using the same instance of the corresponding aggregation, making it shareable in utilizing the weights. An element-wise addition operation sums up the output of both passes and to avoid overfitting, a dropout of 0.1 is applied to the output of each aggregator.

\section{Experimental Validation}
This section provides details of the datasets, implementation, and our experimental results showing our approach's superior performance  compared to the state-of-the-art.

\subsection{Datasets Description} We use the following two datasets for experimental validation of our method. The CMRxsegmentation dataset provides a balanced gender distribution (160 females, 140 males) and age diversity (mean age 26 ± 5 years), ensuring a representative analysis. It also includes multi-contrast CMR images, offering comprehensive coverage of cardiac tissue characteristics. The M\&Ms-2 dataset represents clinically relevant cardiac conditions and offers diverse anatomical variations. Additionally, the dataset includes disease and healthy subjects, making the CAMS-Net robust and generalizable to various clinical scenarios.

\noindent \textbf{CMR×Recon Segmentation data}: The CMR×Recon MICCAI-2023 challenge \cite{wang2023cmrxrecon} data has multi-contrast, multi-view, multi-slice, and multi-coil cardiac magnetic resonance imaging (MRI) data from 300 subjects. The research community uses the data for both reconstruction and segmentation tasks \cite{qayyum2024transforming,zhou2024simultaneous}; in the proposed study, we have only used it as segmentation data. The challenge includes short-axis (SAX), two-chamber (2CH), three-chamber (3CH), four-chamber (4CH) long-axis (LAX) views, and T1 mapping and T2 mappings. We used the 4CH-LAX cine images and corresponding segmentation labels, which have been manually labeled by an expert radiologist where annotations are provided for four Cardiac chambers: LA (label=1), RA (label=2), LV (label=3), and RV (label=4). We utilize a randomly selected five-fold cross-validation split of the CMR×Segmentation dataset. 

\noindent \textbf{M\&Ms-2 data}: The M\&Ms-2 is MICCAI 2021 challenge \cite{campello2021multi,martin2023deep}, focused on RV segmentation and provided labels for LV, RV, and LV-myocardium (MYO).  The data is collected from three clinical centers in Spain utilizing nine scanners from three vendors (Siemens, General Electric, and Philips).  It contains 360 subjects, sequentially divided into 160 for training, 40 for validation, and 160 for testing. In the proposed study, we have used the LA view segmentation images, and similar to \cite{liu2022transfusion}, all models are evaluated using a 5-fold cross-validation split. 

\subsection{Implementation Details} Comparative networks and the proposed framework were implemented using PyTorch, and all experiments were performed using NVidia A100 GPUs with 40GB RAM.
AdamW \cite{loshchilov2018decoupled} optimizer is used with $\beta_1$, $\beta_2$ = [0.5,0.55]; training was performed for 500 training epochs using Dice Loss \cite{jadon2020survey} with an initial learning rate of $1e^{-4}$ which was halved after every 100 epochs. Similar to \cite{liu2024swin}, we pre-trained the encoder part of the proposed CAMS-Net on ImageNet \cite{russakovsky2015imagenet}, followed by fine-tuning it on the segmentation data. We pre-process each input intensity image by normalizing it by its mean and standard deviation. Various intensity and geometric data augmentation are applied to improve the diversity of training, including Gaussian noise, blur, brightness contrast, random ghosting, rotation, scaling, random flipping, and random affine. For a fair comparison, all the networks are trained with the same protocol and fine-tuned where the pre-trained weights were available. 

\begin{table*}[t!]
\centering
        \resizebox{0.9\textwidth}{!}{
        \footnotesize
        \begin{tabular}{|c|c|c|c|c|c|c|c|c|c|c|c|}
            \hline
             \multirow{2}{*}{Methods} & \multirow{2}{*}{ \# Params (M)$\downarrow$} & \multicolumn{5}{|c|}{Dice Score ($\%$) $\uparrow$}  & \multicolumn{5}{|c|}{ Hausdorff Distance- HD (mm) $\downarrow$} \\
             \cline{3-12}
              &  & LA & RA & LV & RV & Avg & LA & RA & LV & RV  & Avg \\ 
            \hline
            UNet\cite{ronneberger2015u}  & {31.03} & {{85.36}} & 79.81 & 90.39 & 85.17 & {{85.18}} & {4.66} & 10.71 & 6.11 & {5.46} & 6.73 \\
            ResUNet\cite{diakogiannis2020resunet} & 46.41   & 82.34 & 79.27 & {91.38} & {85.21} & 84.55 & 5.36 & 6.13 & {{3.78}} & {{5.32}} & {5.14} \\
            UNETR\cite{hatamizadeh2022unetr} & 95.85   & {83.45} & {81.18} & 90.62 & 84.64 & 84.97 & 5.20 & {5.95} & 5.31 & 6.18 & 5.66 \\
            Swin-UNETR\cite{hatamizadeh2021swin} & {25.13}    & 82.70 & {81.33} & 91.19 & 85.14 & {85.09} & 5.50 & {{5.73}} & 4.48 & 5.93 & 5.41 \\
            Swin-UNet\cite{cao2022swin} & 41.35  & 83.48 & {{81.91}} & 90.60 & 84.54 & {85.13} & {{4.51}} & 6.28 & 4.48 & 6.10 & 5.34 \\
            TransUNet\cite{chen2021transunet} & 66.80 & 85.23 & 82.02 & 91.59 & 85.50 & 86.08 & 5.27 & 6.69 & 5.69 & 6.98 & 6.15\\
            VM-UNet\cite{ruan2024vm} & 44.27   & 80.65 & 80.30 & {{91.59}} & {85.22} & 84.44 & {4.74} & {5.90} & {3.98} &  {5.59} & {{5.05}} \\
            Mamba-UNet\cite{wang2024mamba}  & 35.85   & 82.27 & 81.06 & {91.58} & {{85.33}} & 85.05 & 4.95 & 6.29 & {4.02} & 5.73 & {5.24}\\
             \hline
            \textbf{CAMS-Net(ours)} & \textbf{18.56}  & \textbf{86.06} & \textbf{84.44} & \textbf{92.53} & \textbf{87.35} & \textbf{87.59} & \textbf{4.07} & \textbf{5.43} & \textbf{3.10} &  \textbf{5.40} & \textbf{4.50}\\           
            \hline
        \end{tabular}
        }
\caption{Comparison of state-of-the-art methods using a five-fold cross-validation split of CMR-Segmentation Dataset. Best results are shown in \textbf{Bold} and model parameters (\# Params) are listed in millions (M).\label{tab_main}}
\end{table*} 

\begin{figure*}[t!]
\centering
\includegraphics[scale=0.41]{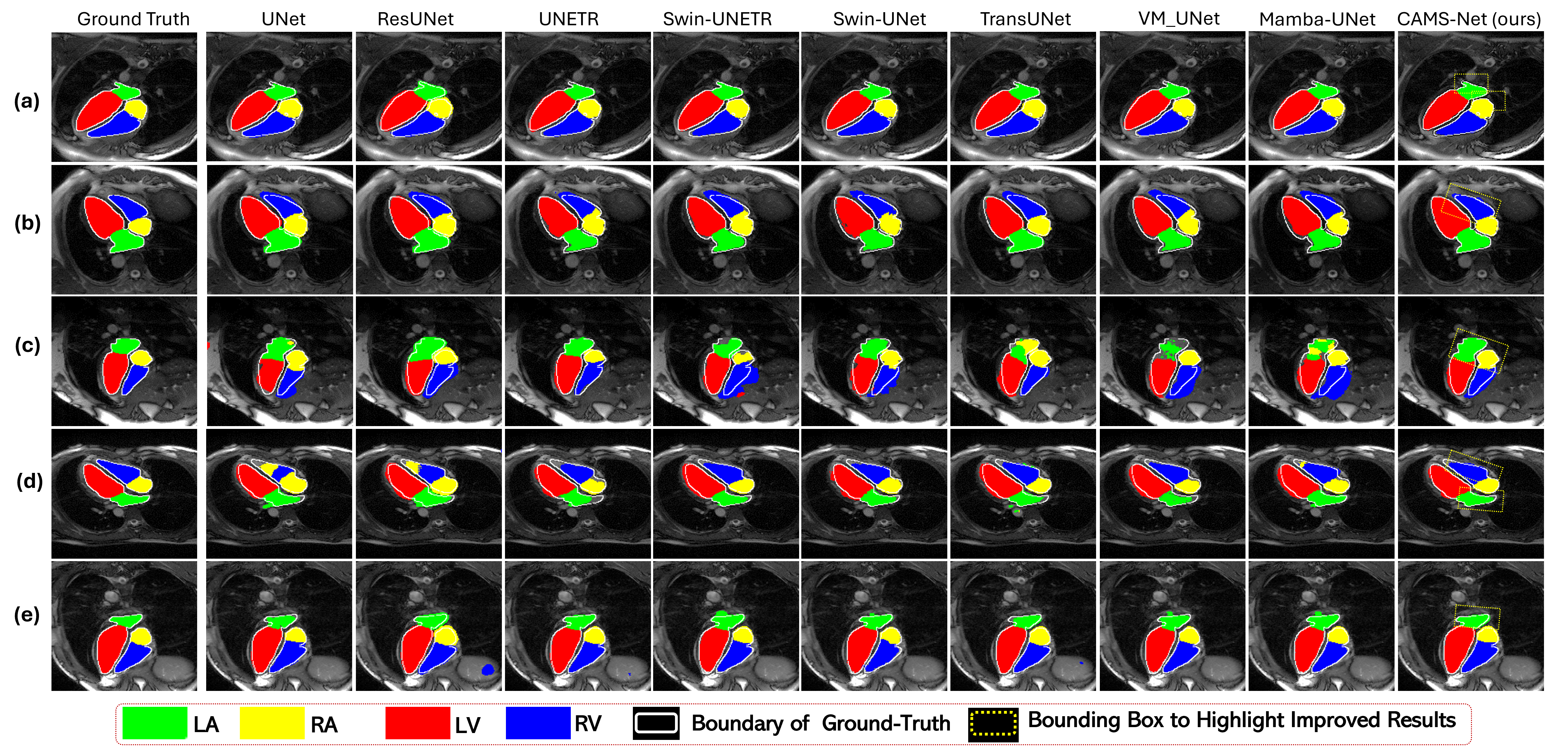}
\caption{Qualitative comparison from CMRxSegmentation dataset, using CAMS-Net and other networks, highlighting CAMS-Net's enhanced performance in boundary separation and preserving spatial integrity across different regions. Please zoom in for details.} 
\label{viz1}
\end{figure*}

\subsection{Experimental Validation with CMR Dataset}
\label{sec:results}
We compared and performed experiments with a number of state-of-the-art methods, including (i) CNNs-based models: UNet\cite{ronneberger2015u} and ResUNet\cite{diakogiannis2020resunet}, (ii) self-attention-based Swin-UNet\cite{cao2022swin}, (iii) hybrid-architectures (CNNs+self-attention): UNETR\cite{hatamizadeh2022unetr} and Swin-UNETR\cite{hatamizadeh2021swin}, and (iv) Mamba-based models: VM-UNet\cite{ruan2024vm} and Mamba-UNet\cite{wang2024mamba}.
\tabref{tab_main} shows experimental validation results using five-fold cross-validation with the CMR-segmentation dataset. The proposed method outperforms existing methods while requiring the least model parameters. 
We note our method has 18.56 million trainable parameters, compared to counterparts with parameter counts ranging from 25.13 to 95.85 million. We attribute this to a combination of architectural innovations, including the LIFM block, which factorizes the Mamba block; the CS-IF module, which includes MCA and MSA capturing information along both channel and spatial dimensions; and the proposed bidirectional scanning scheme augmented with the proposed weight-sharing strategy.

\figref{viz1} shows the visual results of CAMS-Net and each comparative network. For rows (a) and (b), the proposed CAMS-Net column shows precise segmentation of anatomical structures, delineating clear boundaries. Specifically, row (a) shows the accurate segmentation of the LA boundary, while row (b) illustrates the distinct segmentation of the RV and RA boundaries compared to other networks. In row (c), where all other networks fail to reconstruct RA and LA correctly, the proposed CAMS-Net can segment these anatomies precisely with clear boundaries; we attribute this improvement to the incorporation of MSA, which helps maintain spatial coherence, as also demonstrated in ablation studies (also see \figref{abl_ciz}). For the last two rows of \figref{viz1}, where the comparative networks either over-segment or under-segment, the CAMS-Net still performs comparably by ensuring balanced and accurate segmentation. In row (d), the CAMS-Net preserves the shape and structure of RV, compared to UNet, ResUNet, and Mamba-UNet, which segment sections of RV as RA. Finally, in row (e), where ResUNet, UNETR, and TransUNet generate a floating prediction for RV and other networks fail to delineate LA boundaries, the proposed CAMS-Net can gather spatial context from both directions because of Bidirectional scanning, which helps it to resolve ambiguities regions.

\begin{table*}[t!]
\centering
        \resizebox{0.68\textwidth}{!}{
        \begin{tabular}{|c|c|c|c|c|c|c|c|c|}
            \hline
             \multirow{2}{*}{Methods} &  \multicolumn{4}{|c|}{Dice Score ($\%$) $\uparrow$} & \multicolumn{4}{|c|}{HD (mm) $\downarrow$}\\
             \cline{2-9}
              & LV & RV & Myo & Avg & LV & RV & Myo & Avg\\ 
            \hline
            UNet\cite{ronneberger2015u}    & 87.26 & 88.20 & 79.96 & 85.14 & 13.04 & 8.76 & 12.24 & 11.35\\
            ResUNet\cite{diakogiannis2020resunet}   & 87.61 & 88.41 & 80.12 & 85.38 & 12.72 & 8.39 & 11.28 & 10.80\\
            InfoTrans*\cite{li2021right} &  88.21 & 89.11 & 80.55 & 85.96 & 12.47 & 7.23 & 10.21 & 9.97\\
            TransUNet\cite{chen2021transunet}  & 87.91 & 88.23 & 79.05 & 85.06 & 12.02 & 8.14 & 11.21 & 10.46\\
            MCTrans\cite{ji2021multi}   & 88.42 & 88.19 & 79.47 & 85.36 & 11.78 & 7.65 & 10.76 & 10.06\\
            MCTrans*\cite{ji2021multi}  &  88.81 & 88.61 & 79.94 & 85.79 & 11.52 & 7.02 & 10.07 & 9.54\\
            UTNet\cite{gao2021utnet} &  86.93 & 89.07 & 80.48 & 85.49 & 11.47 & 6.35 & 10.02 & 9.28\\
            UTNet*\cite{gao2021utnet} &  87.36 & {90.42} & 81.02 & 86.27 & 11.13 & {5.91} & {9.81} & {8.95}\\
            SWIN-UNET\cite{cao2022swin}   & {90.88} & {86.66} & {79.93} & {85.82} & {10.08} & {10.08} & {6.07} & {8.74}\\
            UNETR\cite{hatamizadeh2022unetr}   & {91.08} & {87.30} & {81.17} & {86.52} & {8.86} & {9.93} & {5.98} & {8.25}\\
            SWIN-UNETR\cite{hatamizadeh2021swin}   & {91.91} & {86.77} & {82.36} & {87.01} & {7.19} & {8.77} & {4.65} & {6.87}\\
            TransFusion*\cite{liu2022transfusion}   & {89.78} & {91.52} & {81.79} & {88.70} & {10.25} & {5.12} & {8.69} & {8.02}\\
             VM-UNet\cite{ruan2024vm} & 92.47 &  87.79 & 82.39 &  87.55 & 6.09 &  7.60 & 4.87 &  6.18  \\
              Mamba-UNet\cite{wang2024mamba}  & 93.44 & 87.18 & 82.54 &  87.72 & 6.63 & 8.08 & 5.59 & 6.76\\
             \hline
             \textbf{CAMS-Net (ours)} & \textbf{94.45} &  \textbf{91.87} &  \textbf{86.21} &  \textbf{90.84} &  \textbf{3.64} &  \textbf{4.79} &  \textbf{2.61} &  \textbf{4.34}\\
             \hline
        \end{tabular}}
\caption{Comparison of results obtained from different methods using a five-fold cross-validation split of M\&Ms-2 dataset.
Methods indicated with a $*$ use multi-view inputs. Best results are shown in \textbf{Bold}.}\label{tabMnM2}
\end{table*}
\begin{figure*}[ht!]
\centering
\includegraphics[scale=0.36]{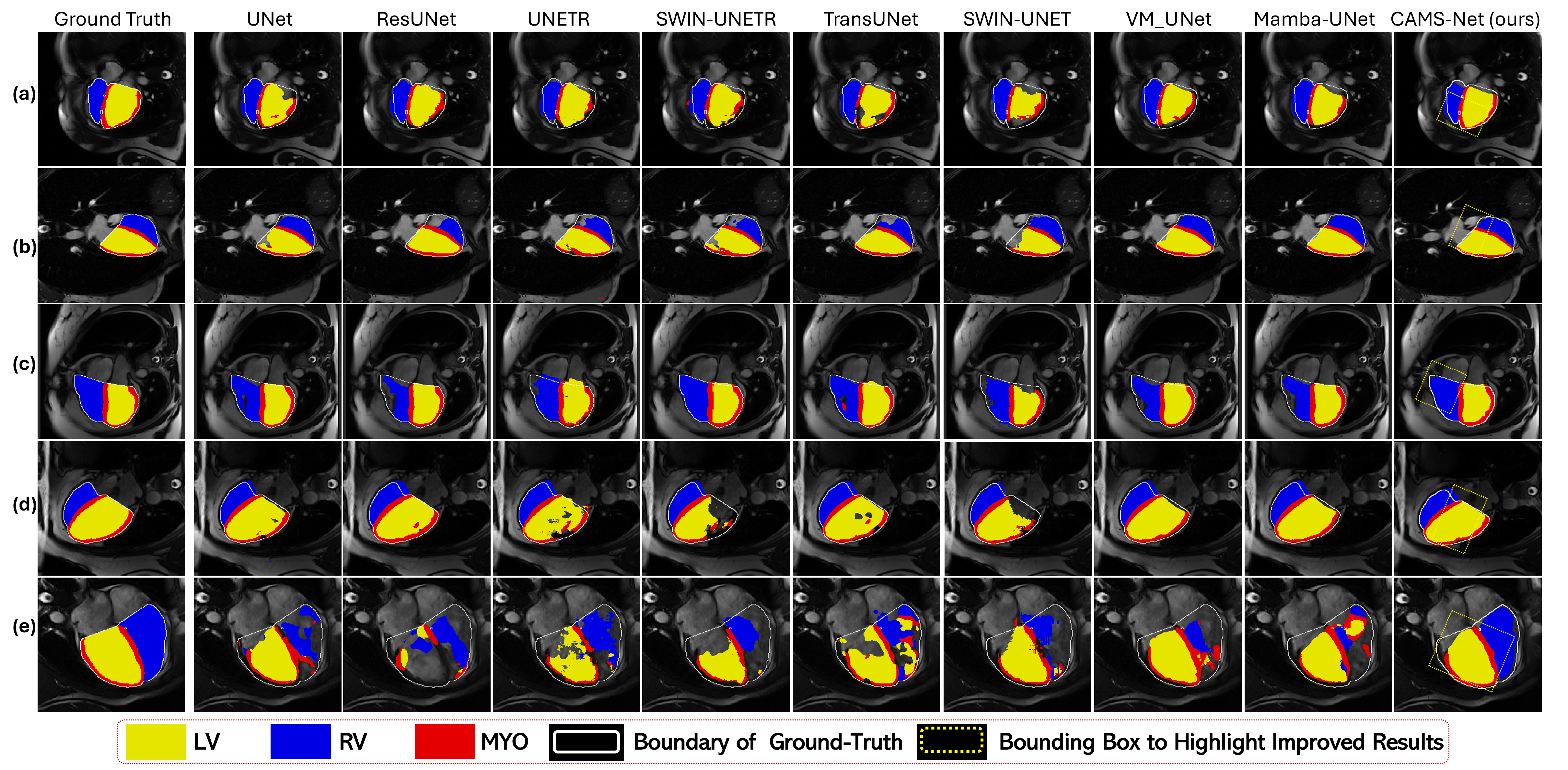}
\caption{Qualitative comparison of visual results of CAMS-Net and other networks using M\&Ms2 dataset. Please zoom in for details.} 
\label{mnm_viz}
\end{figure*}

\subsection{Experimental Validation with M\&Ms-2 Dataset}
\label{MnM2:abl}
To show the capability of CAMS-Net to work on multiple datasets, we also perform additional experimental validation using the M\&Ms-2 dataset. \tabref{tabMnM2} shows the CAMS-Net performance compared to existing methods, where it achieves the highest Dice Score across all categories and the lowest HD values, highlighting CAMS-Net's superior boundary accuracy.

The visual results shown in \figref{mnm_viz} further advocate the superior performance of the proposed CAMS-Net. In rows (a) and (b), all networks struggle to differentiate between the LV and MYO's boundaries and fail to accurately capture the variable shapes of RV, compared to the proposed CAMS-Net's results, where it generates clear boundaries. In row (c), the other comparative networks, except SWIN-UNETR, are unable to segment the RV properly, resulting in higher false negatives. In row (d), except for the Mamba-based networks, all other comparative networks cannot capture the relationship between MYO and LV. Our CAMS-Net's predictions closely match the ground truth, confirming its capability of capturing long-range dependencies. For the last row (e), all other networks produce incomplete segmentations compared to the CAMS-Net, demonstrating spatial continuity and coherence in segmenting all three regions of interest, which is attributed to our proposed MSA module that captures information along spatial dimensions (see \figref{abl_ciz} and Section \ref{sec:abl} for further analysis).

\section{Ablation Studies}
\label{sec:abl}
We performed the following ablation studies on the CMR dataset to show how each proposed module contributes to improved accuracy. 
\begin{table*}[t!]
\centering
\resizebox{0.8\textwidth}{!}{
\footnotesize
\begin{tabular}{|c|c|c|c|c|c|c|c|c|c|c|}
\hline
\cline{7-11}
\multicolumn{1}{|c|}{\shortstack[c]{MCA}} & \multicolumn{1}{c|}{\shortstack[c]{MSA}} & \multicolumn{1}{c|}{\shortstack[c]{Bidirectional \\ Scanning}} & \multicolumn{1}{c|}{\shortstack[c]{Weights\\Sharing }} & \multicolumn{1}{c|}{\shortstack[c]{Positional \\ Embeddings}} & \multicolumn{1}{c|}{\shortstack[c]{Pretraining}} & LA & RA & LV & RV & Avg \\
\hline
 \cmark  & \xmark & \xmark & \xmark & \xmark & \xmark & 80.15 & 78.33 & 89.65 & 83.22 & 82.83 \\
  \cmark  & \cmark & \xmark & \xmark & \xmark & \xmark & {82.17} & {81.09} & {90.30} & {83.79} & {84.33} \\
  \hline
 \cmark  & \cmark & \cmark & \xmark & \xmark & \xmark & 83.27 & 81.83 & 90.27 & 84.76 & 85.03 \\
 \cmark  & \cmark & \cmark & \cmark & \xmark & \xmark &  {83.20} & {81.10} & {91.46} & {84.68} & {85.11} \\
 \hline
  \cmark  & \cmark & \cmark & \cmark & \cmark & \xmark & {84.11} & {82.09} & {91.50} & {84.51} & {85.55} \\
 \cmark  & \cmark & \cmark & \cmark & \cmark & \cmark & \textbf{86.06} & \textbf{84.44} & \textbf{92.53} & \textbf{87.35} & \textbf{87.59} \\
\hline
\end{tabular}
}
\caption{Ablation studies (Dice score \%)  utilizing MCA, MSA, bidirectional scanning augmented with weight sharing strategy, positional embeddings, and pretraining on ImageNet using a five-fold cross-validation split of CMR-segmentation dataset.\\}\label{tab1}
\end{table*}
\begin{figure}[ht!]
\centering
\includegraphics[scale=0.27]{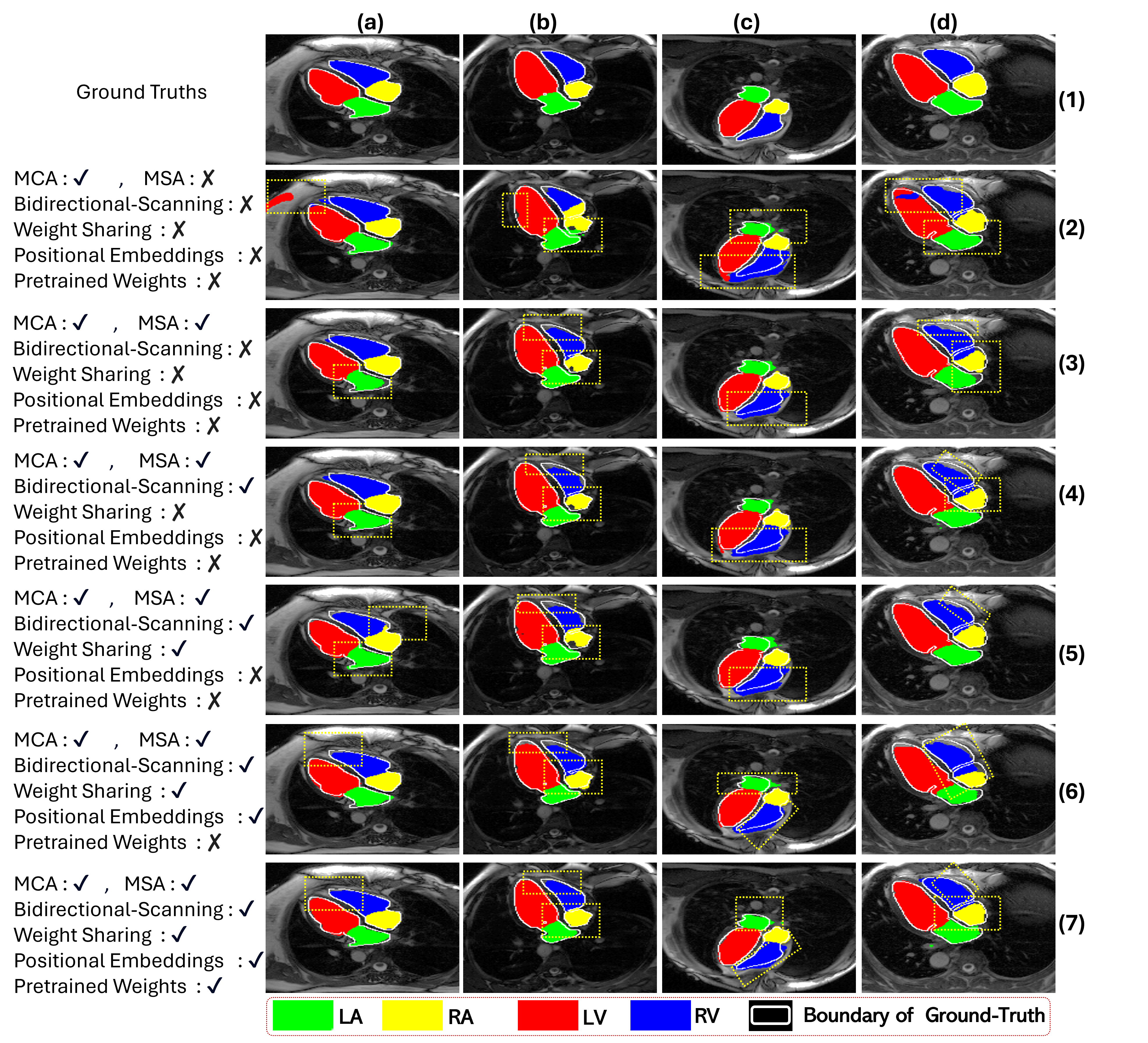}
\caption{Visual comparison of results from different ablation studies on CMR-segmentation dataset. Please zoom in for details.} 
\label{abl_ciz}
\end{figure}

\noindent\textbf{CAMS-Net W/ and WO/ MSA:} The MSA fosters intercommunication among spatial locations, and its effectiveness is evaluated by removing it from the CS-IF module and using MCA throughout the network. The first two rows of \tabref{tab1} list the quantitative results of utilizing MSA and MCA, bringing an average improvement of 1.5\% in the Dice score. Also, shown in the third row of \figref{abl_ciz}, the MSA enables the model to learn spatial dependencies between different regions of the image, resulting in better delineation of boundaries, better spatial coherence, fewer errors in spatial relationships, and generally improved localization of anatomical structures.  
Note that we have utilized the default scanning strategy from Mamba \cite{gu2023mamba}.

\noindent\textbf{Bidirectional scanning and weight sharing strategy:} The bidirectional scanning scheme incorporated at each encoder-decoder stage improves the results, shown in row 3 of the \tabref{tab1}. However, this comes at the cost of extra parameters for each forward and backward scanning scheme. The proposed bidirectional scanning, augmented with a sharing strategy, overcomes this limitation by reducing the parameters and improving the overall performance, shown in row 4 of \tabref{tab1}. Rows 4 and 5 of \figref{abl_ciz} depict how bidirectional scanning improves the results by capturing relationships from both directions, specifically in column (d); it helps the network better delineate the boundary between RV and RA.

\noindent\textbf{Positional embeddings:} CAMS-Net also utilizes the sinusoidal positional embeddings to encode spatial information about the position of each element within the input image sequence \cite{vaswani2017attention}. Row 5 of \tabref{tab1} lists the results of utilizing positional embeddings, which helps the network improve its average accuracy. Columns (a) and (b) of row 6 in \figref{abl_ciz} exhibit how it can maintain spatial consistency across different regions, i.e., for RA here.


\noindent\textbf{Effects of pre-training:} We conducted experiments to examine the impact of ImageNet pre-trained parameters on the proposed CAMS-Net’s performance. The last row of \tabref{tab1} lists the experimental results of this ablation and shows that utilizing the pre-trained weights improves the average Dice score by > 2\%. Row 7 of \tabref{tab1} shows how pre-training leverages visual features like edges, textures, and shapes it has learned from ImageNet and improves the CAMS-Net’s ability to detect boundaries more accurately.


\section{Conclusion and Future Work}
\label{sec:conc}
We are the first to propose a Mamba-based segmentation network without convolution operations and self-attention mechanisms to showcase the power of SSM-based architectures. We introduced several innovative strategies to the Mamba-based methods to increase their performance and reduce the computational complexity, including (i) a Linearly Interconnected Factorized Mamba (LIFM) block to reduce the number of trainable parameters and increase decision function,  (ii) Mamba-based channel and spatial aggregators to learn the information across different channels along with spatial locations of the features, and (iii) a bidirectional weight-sharing strategy scheme. Our experiments demonstrate that the proposed CAMS-Net, an SSM-based segmentation network, outperforms the existing state-of-the-art in CNN, self-attention, and Mamba-based methods on CMR and M\&Ms-2 segmentation datasets. 

CAMS-Net has been implemented for 2D medical image segmentation, which has demonstrated impressive results in segmenting anatomical structures from medical scans like cardiac MRIs. However, there is significant potential to extend this work to 3D medical image segmentation, which we will explore in our future work. 
 
\newpage
{\small
\bibliographystyle{ieee_fullname}
\bibliography{egbib}

\begin{thebibliography}{10}\itemsep=-1pt

\bibitem{aljuaid2022survey}
Abeer Aljuaid and Mohd Anwar.
\newblock Survey of supervised learning for medical image processing.
\newblock {\em SN Computer Science}, 3(4):292, 2022.

\bibitem{campello2021multi}
Victor~M Campello, Polyxeni Gkontra, Cristian Izquierdo, Carlos Martin-Isla, Alireza Sojoudi, Peter~M Full, Klaus Maier-Hein, Yao Zhang, Zhiqiang He, Jun Ma, et~al.
\newblock Multi-centre, multi-vendor and multi-disease cardiac segmentation: the m\&ms challenge.
\newblock {\em IEEE Transactions on Medical Imaging}, 40(12):3543--3554, 2021.

\bibitem{cao2022swin}
Hu Cao, Yueyue Wang, Joy Chen, Dongsheng Jiang, Xiaopeng Zhang, Qi Tian, and Manning Wang.
\newblock Swin-unet: Unet-like pure transformer for medical image segmentation.
\newblock In {\em European conference on computer vision}, pages 205--218. Springer, 2022.

\bibitem{chen2020deep}
Chen Chen, Chen Qin, Huaqi Qiu, Giacomo Tarroni, Jinming Duan, Wenjia Bai, and Daniel Rueckert.
\newblock Deep learning for cardiac image segmentation: a review.
\newblock {\em Frontiers in cardiovascular medicine}, 7:25, 2020.

\bibitem{chen2021transunet}
Jieneng Chen, Yongyi Lu, Qihang Yu, Xiangde Luo, Ehsan Adeli, Yan Wang, Le Lu, Alan~L Yuille, and Yuyin Zhou.
\newblock Transunet: Transformers make strong encoders for medical image segmentation.
\newblock {\em arXiv preprint arXiv:2102.04306}, 2021.

\bibitem{dao2023hungry}
Tri Dao, Daniel~Y Fu, Khaled~K Saab, Armin~W Thomas, Atri Rudra, and Christopher R{\'e}.
\newblock Hungry hungry hippos: Towards language modeling with state space models.
\newblock In {\em Proceedings of the 11th International Conference on Learning Representations (ICLR)}, 2023.

\bibitem{diakogiannis2020resunet}
Foivos~I Diakogiannis, Fran{\c{c}}ois Waldner, Peter Caccetta, and Chen Wu.
\newblock Resunet-a: A deep learning framework for semantic segmentation of remotely sensed data.
\newblock {\em ISPRS Journal of Photogrammetry and Remote Sensing}, 162:94--114, 2020.

\bibitem{dosovitskiy2020image}
Alexey Dosovitskiy, Lucas Beyer, Alexander Kolesnikov, Dirk Weissenborn, Xiaohua Zhai, Thomas Unterthiner, Mostafa Dehghani, Matthias Minderer, Georg Heigold, Sylvain Gelly, et~al.
\newblock An image is worth 16x16 words: Transformers for image recognition at scale.
\newblock In {\em International Conference on Learning Representations}, 2020.

\bibitem{drozdzal2016importance}
Michal Drozdzal, Eugene Vorontsov, Gabriel Chartrand, Samuel Kadoury, and Chris Pal.
\newblock The importance of skip connections in biomedical image segmentation.
\newblock In {\em International workshop on deep learning in medical image analysis, international workshop on large-scale annotation of biomedical data and expert label synthesis}, pages 179--187. Springer, 2016.

\bibitem{elfwing2018sigmoid}
Stefan Elfwing, Eiji Uchibe, and Kenji Doya.
\newblock Sigmoid-weighted linear units for neural network function approximation in reinforcement learning.
\newblock {\em Neural networks}, 107:3--11, 2018.

\bibitem{gao2021utnet}
Yunhe Gao, Mu Zhou, and Dimitris~N Metaxas.
\newblock Utnet: a hybrid transformer architecture for medical image segmentation.
\newblock In {\em Medical Image Computing and Computer Assisted Intervention--MICCAI 2021: 24th International Conference, Strasbourg, France, September 27--October 1, 2021, Proceedings, Part III 24}, pages 61--71. Springer, 2021.

\bibitem{geirhos2018imagenet}
Robert Geirhos, Patricia Rubisch, Claudio Michaelis, Matthias Bethge, Felix~A Wichmann, and Wieland Brendel.
\newblock Imagenet-trained cnns are biased towards texture; increasing shape bias improves accuracy and robustness.
\newblock In {\em International Conference on Learning Representations}, 2018.

\bibitem{gu2023mamba}
Albert Gu and Tri Dao.
\newblock Mamba: Linear-time sequence modeling with selective state spaces.
\newblock {\em arXiv preprint arXiv:2312.00752}, 2023.

\bibitem{gu2021efficiently}
Albert Gu, Karan Goel, and Christopher Re.
\newblock Efficiently modeling long sequences with structured state spaces.
\newblock In {\em International Conference on Learning Representations}, 2021.

\bibitem{hatamizadeh2021swin}
Ali Hatamizadeh, Vishwesh Nath, Yucheng Tang, Dong Yang, Holger~R Roth, and Daguang Xu.
\newblock Swin unetr: Swin transformers for semantic segmentation of brain tumors in mri images.
\newblock In {\em International MICCAI Brainlesion Workshop}, pages 272--284. Springer, 2021.

\bibitem{hatamizadeh2022unetr}
Ali Hatamizadeh, Yucheng Tang, Vishwesh Nath, Dong Yang, Andriy Myronenko, Bennett Landman, Holger~R Roth, and Daguang Xu.
\newblock Unetr: Transformers for 3d medical image segmentation.
\newblock In {\em Proceedings of the IEEE/CVF winter conference on applications of computer vision}, pages 574--584, 2022.

\bibitem{hesamian2019deep}
Mohammad~Hesam Hesamian, Wenjing Jia, Xiangjian He, and Paul Kennedy.
\newblock Deep learning techniques for medical image segmentation: achievements and challenges.
\newblock {\em Journal of digital imaging}, 32:582--596, 2019.

\bibitem{hu2024zigma}
Vincent~Tao Hu, Stefan~Andreas Baumann, Ming Gui, Olga Grebenkova, Pingchuan Ma, Johannes Fischer, and Bjorn Ommer.
\newblock Zigma: Zigzag mamba diffusion model.
\newblock {\em arXiv preprint arXiv:2403.13802}, 2024.

\bibitem{jadon2020survey}
Shruti Jadon.
\newblock A survey of loss functions for semantic segmentation.
\newblock In {\em 2020 IEEE conference on computational intelligence in bioinformatics and computational biology (CIBCB)}, pages 1--7. IEEE, 2020.

\bibitem{ji2021multi}
Yuanfeng Ji, Ruimao Zhang, Huijie Wang, Zhen Li, Lingyun Wu, Shaoting Zhang, and Ping Luo.
\newblock Multi-compound transformer for accurate biomedical image segmentation.
\newblock In {\em MICCAI}, pages 326--336. Springer, 2021.

\bibitem{karimi2021convolution}
Davood Karimi, Serge~Didenko Vasylechko, and Ali Gholipour.
\newblock Convolution-free medical image segmentation using transformers.
\newblock In {\em Medical Image Computing and Computer Assisted Intervention--MICCAI 2021: 24th International Conference, Strasbourg, France, September 27--October 1, 2021, Proceedings, Part I 24}, pages 78--88. Springer, 2021.

\bibitem{khan2024crop}
Abbas Khan, Muhammad Asad, Martin Benning, Caroline Roney, and Gregory Slabaugh.
\newblock Crop and couple: cardiac image segmentation using interlinked specialist networks.
\newblock {\em arXiv e-prints}, pages arXiv--2402, 2024.

\bibitem{khan2023transformers}
Rabeea~Fatma Khan, Byoung-Dai Lee, and Mu~Sook Lee.
\newblock Transformers in medical image segmentation: a narrative review.
\newblock {\em Quantitative Imaging in Medicine and Surgery}, 13(12):8747, 2023.

\bibitem{kim2022restr}
Namyup Kim, Dongwon Kim, Cuiling Lan, Wenjun Zeng, and Suha Kwak.
\newblock Restr: Convolution-free referring image segmentation using transformers.
\newblock In {\em Proceedings of IEEE/CVF Conference on Computer Vision and Pattern Recognition (CVPR)}, June 2022.

\bibitem{li2021right}
Lei Li, Wangbin Ding, Liqin Huang, and Xiahai Zhuang.
\newblock Right ventricular segmentation from short-and long-axis mris via information transition.
\newblock In {\em STACOM}, pages 259--267. Springer, 2021.

\bibitem{litjens2019state}
Geert Litjens, Francesco Ciompi, Jelmer~M Wolterink, Bob~D de Vos, Tim Leiner, Jonas Teuwen, and Ivana I{\v{s}}gum.
\newblock State-of-the-art deep learning in cardiovascular image analysis.
\newblock {\em JACC: Cardiovascular imaging}, 12(8 Part 1):1549--1565, 2019.

\bibitem{liu2022transfusion}
Di Liu, Yunhe Gao, Qilong Zhangli, Ligong Han, Xiaoxiao He, Zhaoyang Xia, Song Wen, Qi Chang, Zhennan Yan, Mu Zhou, et~al.
\newblock Transfusion: multi-view divergent fusion for medical image segmentation with transformers.
\newblock In {\em MICCAI}, pages 485--495. Springer, 2022.

\bibitem{liu2024swin}
Jiarun Liu, Hao Yang, Hong-Yu Zhou, Yan Xi, Lequan Yu, Yizhou Yu, Yong Liang, Guangming Shi, Shaoting Zhang, Hairong Zheng, et~al.
\newblock Swin-umamba: Mamba-based unet with imagenet-based pretraining.
\newblock {\em arXiv preprint arXiv:2402.03302}, 2024.

\bibitem{liu2024vmamba}
Yue Liu, Yunjie Tian, Yuzhong Zhao, Hongtian Yu, Lingxi Xie, Yaowei Wang, Qixiang Ye, and Yunfan Liu.
\newblock Vmamba: Visual state space model.
\newblock {\em arXiv preprint arXiv:2401.10166}, 2024.

\bibitem{liu2021swin}
Ze Liu, Yutong Lin, Yue Cao, Han Hu, Yixuan Wei, Zheng Zhang, Stephen Lin, and Baining Guo.
\newblock Swin transformer: Hierarchical vision transformer using shifted windows.
\newblock In {\em Proceedings of the IEEE/CVF international conference on computer vision}, pages 10012--10022, 2021.

\bibitem{loshchilov2018decoupled}
Ilya Loshchilov and Frank Hutter.
\newblock Decoupled weight decay regularization.
\newblock In {\em International Conference on Learning Representations}, 2018.

\bibitem{martin2023deep}
Carlos Mart{\'\i}n-Isla, V{\'\i}ctor~M Campello, Cristian Izquierdo, Kaisar Kushibar, Carla Sendra-Balcells, Polyxeni Gkontra, Alireza Sojoudi, Mitchell~J Fulton, Tewodros~Weldebirhan Arega, Kumaradevan Punithakumar, et~al.
\newblock Deep learning segmentation of the right ventricle in cardiac mri: The m\&ms challenge.
\newblock {\em IEEE Journal of Biomedical and Health Informatics}, 2023.

\bibitem{martin2020image}
Carlos Martin-Isla, Victor~M Campello, Cristian Izquierdo, Zahra Raisi-Estabragh, Bettina Bae{\ss}ler, Steffen~E Petersen, and Karim Lekadir.
\newblock Image-based cardiac diagnosis with machine learning: a review.
\newblock {\em Frontiers in cardiovascular medicine}, 7:1, 2020.

\bibitem{mehta2022long}
Harsh Mehta, Ankit Gupta, Ashok Cutkosky, and Behnam Neyshabur.
\newblock Long range language modeling via gated state spaces.
\newblock In {\em The Eleventh International Conference on Learning Representations}, 2022.

\bibitem{mehta2022separable}
Sachin Mehta and Mohammad Rastegari.
\newblock Separable self-attention for mobile vision transformers.
\newblock {\em Transactions on Machine Learning Research}, 2022.

\bibitem{niu2021review}
Zhaoyang Niu, Guoqiang Zhong, and Hui Yu.
\newblock A review on the attention mechanism of deep learning.
\newblock {\em Neurocomputing}, 452:48--62, 2021.

\bibitem{patro2024mamba}
Badri~Narayana Patro and Vijay~Srinivas Agneeswaran.
\newblock Mamba-360: Survey of state space models as transformer alternative for long sequence modelling: Methods, applications, and challenges.
\newblock {\em arXiv preprint arXiv:2404.16112}, 2024.

\bibitem{qayyum2024transforming}
Abdul Qayyum, Hao Xu, Brian~P Halliday, Cristobal Rodero, Christopher~W Lanyon, Richard~D Wilkinson, and Steven~Alexander Niederer.
\newblock Transforming heart chamber imaging: Self-supervised learning for whole heart reconstruction and segmentation.
\newblock {\em arXiv preprint arXiv:2406.06643}, 2024.

\bibitem{ramachandran2019stand}
Prajit Ramachandran, Niki Parmar, Ashish Vaswani, Irwan Bello, Anselm Levskaya, and Jon Shlens.
\newblock Stand-alone self-attention in vision models.
\newblock {\em Advances in neural information processing systems}, 32, 2019.

\bibitem{richter2021should}
Mats~L Richter, Julius Sch{\"o}ning, Anna Wiedenroth, and Ulf Krumnack.
\newblock Should you go deeper? optimizing convolutional neural network architectures without training.
\newblock In {\em 2021 20th IEEE International Conference on Machine Learning and Applications (ICMLA)}, pages 964--971. IEEE, 2021.

\bibitem{ronneberger2015u}
Olaf Ronneberger, Philipp Fischer, and Thomas Brox.
\newblock U-net: Convolutional networks for biomedical image segmentation.
\newblock In {\em Medical image computing and computer-assisted intervention--MICCAI 2015: 18th international conference, Munich, Germany, October 5-9, 2015, proceedings, part III 18}, pages 234--241. Springer, 2015.

\bibitem{ruan2024vm}
Jiacheng Ruan and Suncheng Xiang.
\newblock Vm-unet: Vision mamba unet for medical image segmentation.
\newblock {\em arXiv preprint arXiv:2402.02491}, 2024.

\bibitem{russakovsky2015imagenet}
Olga Russakovsky, Jia Deng, Hao Su, Jonathan Krause, Sanjeev Satheesh, Sean Ma, Zhiheng Huang, Andrej Karpathy, Aditya Khosla, Michael Bernstein, et~al.
\newblock Imagenet large scale visual recognition challenge.
\newblock {\em International journal of computer vision}, 115:211--252, 2015.

\bibitem{shaker2023swiftformer}
Abdelrahman Shaker, Muhammad Maaz, Hanoona Rasheed, Salman Khan, Ming-Hsuan Yang, and Fahad~Shahbaz Khan.
\newblock Swiftformer: Efficient additive attention for transformer-based real-time mobile vision applications.
\newblock In {\em Proceedings of the IEEE/CVF International Conference on Computer Vision}, pages 17425--17436, 2023.

\bibitem{simonyan2014very}
Karen Simonyan and Andrew Zisserman.
\newblock Very deep convolutional networks for large-scale image recognition.
\newblock {\em International Conference on Learning Representations}, 2015.

\bibitem{tang2024htc}
Hui Tang, Yuanbin Chen, Tao Wang, Yuanbo Zhou, Longxuan Zhao, Qinquan Gao, Min Du, Tao Tan, Xinlin Zhang, and Tong Tong.
\newblock Htc-net: A hybrid cnn-transformer framework for medical image segmentation.
\newblock {\em Biomedical Signal Processing and Control}, 88:105605, 2024.

\bibitem{tolstikhin2021mlp}
Ilya~O Tolstikhin, Neil Houlsby, Alexander Kolesnikov, Lucas Beyer, Xiaohua Zhai, Thomas Unterthiner, Jessica Yung, Andreas Steiner, Daniel Keysers, Jakob Uszkoreit, et~al.
\newblock Mlp-mixer: An all-mlp architecture for vision.
\newblock {\em Advances in neural information processing systems}, 34:24261--24272, 2021.

\bibitem{vaswani2017attention}
Ashish Vaswani, Noam Shazeer, Niki Parmar, Jakob Uszkoreit, Llion Jones, Aidan~N Gomez, {\L}ukasz Kaiser, and Illia Polosukhin.
\newblock Attention is all you need.
\newblock {\em Advances in neural information processing systems}, 30, 2017.

\bibitem{wang2023cmrxrecon}
Chengyan Wang, Jun Lyu, Shuo Wang, Chen Qin, Kunyuan Guo, Xinyu Zhang, Xiaotong Yu, Yan Li, Fanwen Wang, Jianhua Jin, et~al.
\newblock Cmrxrecon: an open cardiac mri dataset for the competition of accelerated image reconstruction.
\newblock {\em arXiv preprint arXiv:2309.10836}, 2023.

\bibitem{wang2024mamba}
Ziyang Wang, Jian-Qing Zheng, Yichi Zhang, Ge Cui, and Lei Li.
\newblock Mamba-unet: Unet-like pure visual mamba for medical image segmentation.
\newblock {\em arXiv preprint arXiv:2402.05079}, 2024.

\bibitem{yao2024cnn}
Wenjian Yao, Jiajun Bai, Wei Liao, Yuheng Chen, Mengjuan Liu, and Yao Xie.
\newblock From cnn to transformer: A review of medical image segmentation models.
\newblock {\em Journal of Imaging Informatics in Medicine}, pages 1--19, 2024.

\bibitem{zhang2024survey}
Hanwei Zhang, Ying Zhu, Dan Wang, Lijun Zhang, Tianxiang Chen, and Zi Ye.
\newblock A survey on visual mamba.
\newblock {\em arXiv preprint arXiv:2404.15956}, 2024.

\bibitem{zhou2024simultaneous}
Yirong Zhou, Chengyan Wang, Mengtian Lu, Kunyuan Guo, Zi Wang, Dan Ruan, Rui Guo, Peijun Zhao, Jianhua Wang, Naiming Wu, et~al.
\newblock Simultaneous deep learning of myocardium segmentation and t2 quantification for acute myocardial infarction mri.
\newblock {\em arXiv preprint arXiv:2405.10570}, 2024.

\bibitem{zhou2019unet++}
Zongwei Zhou, Md~Mahfuzur~Rahman Siddiquee, Nima Tajbakhsh, and Jianming Liang.
\newblock Unet++: Redesigning skip connections to exploit multiscale features in image segmentation.
\newblock {\em IEEE transactions on medical imaging}, 39(6):1856--1867, 2019.

\bibitem{zhu2024vision}
Lianghui Zhu, Bencheng Liao, Qian Zhang, Xinlong Wang, Wenyu Liu, and Xinggang Wang.
\newblock Vision mamba: Efficient visual representation learning with bidirectional state space model.
\newblock {\em arXiv preprint arXiv:2401.09417}, 2024.

\end{thebibliography}
}

\end{document}